\documentclass{ws-procs11x85}
\usepackage{ws-procs-thm}           
\usepackage{longtable}
\usepackage{subfiles}
\usepackage{booktabs}
\usepackage{url}
\usepackage{wrapfig}
\usepackage{listings}
\usepackage{caption}
\begin{document}

\title{Local Large Language Models for Complex Structured Tasks}

\author{V. K. Cody Bumgardner$^\dag$$^1$$^2$, Aaron Mullen$^2$, Sam Armstrong$^2$, Caylin Hickey$^3$, Jeff Talbert$^2$}

\address{
$^1$Department of Pathology and Laboratory Medicine\\
$^2$Institute for Biomedical Informatics\\
$^3$UKHC Clinical Laboratory \\
University of Kentucky\\
Lexington, Kentucky 40506, USA\\
$^\dag$E-mail: ai@uky.edu\\
medicine.ai.uky.edu}

\begin{abstract}
This paper introduces an approach that combines the language reasoning capabilities of large language models (LLMs) with the benefits of local training to tackle complex, domain-specific tasks. Specifically, the authors demonstrate their approach by extracting structured condition codes from pathology reports. The proposed approach utilizes local LLMs, which can be fine-tuned to respond to specific generative instructions and provide structured outputs.  The authors collected a dataset of over 150k uncurated surgical pathology reports, containing gross descriptions, final diagnoses, and condition codes.  They trained different model architectures, including LLaMA, BERT and LongFormer and evaluated their performance. The results show that the LLaMA-based models significantly outperform BERT-style models across all evaluated metrics, even with extremely reduced precision.  The LLaMA models performed especially well with large datasets, demonstrating their ability to handle complex, multi-label tasks.  Overall, this work presents an effective approach for utilizing LLMs to perform domain-specific tasks using accessible hardware, with potential applications in the medical domain, where complex data extraction and classification are required.
\end{abstract}



\copyrightinfo{Preprint of an article submitted for consideration in Pacific Symposium on Biocomputing \copyright\ 2024 copyright World Scientific Publishing Company https://www.worldscientific.com/}


\lstdefinestyle{codelistingstyle}{
  captionpos=b,
  stepnumber=1,
  numbersep=6pt,
  tabsize=4,
}
\lstset{style=codelistingstyle}

\section{Introduction}
\label{introduction}

In recent years, artificial intelligence (AI) and natural language processing (NLP) have been applied to medicine from clinical prognosis to diagnostic and companion diagnostic services. One of the most potentially groundbreaking developments in this domain has been the emergence of generative large language models (LLM), such as OpenAI's ChatGPT\cite{openaichatgpt} and its successors. These user-facing AI-driven systems have proven to be attractive resources, revolutionizing the way medical professionals interact with AI for both research and patient care.

LLMs possess a great capacity to analyze vast amounts of medical data, ranging from research papers and clinical trial results to electronic health records and patient narratives\cite{xue2023potential, dave2023chatgpt}.  By integrating these diverse, potentially multimodal\cite{li2023llava} data sources, these models can identify patterns, correlations, and insights that might have otherwise remained hidden.  With their ability to understand natural language, these AI-powered systems can process patient symptoms, medical histories, and test results to aid in diagnosing diseases more efficiently.  LLMs have demonstrated encouraging capabilities to generate \cite{zhou2023evaluation} and summarize \cite{temsah2023overview} medical reports, including radiology \cite{ma2023impressiongpt, biswas2023chatgpt, jeblick2022chatgpt} and pathology \cite{pathchatgpt, sinha2023applicability, brennan2023using} diagnostic reports.

The large volume of language data used in training LLMs has enabled so-called zero-shot \cite{wang2019survey} data operations across classes of data not necessarily observed during model training.  While LLMs are useful for many transferable language tasks, performance is dependent on distinguishable associations between observed and non-observed classes.  Medical terminologies, domain specific-jargon, and institutional reporting practices produce unstructured data that does not necessarily contain transferable associations used by general-purpose LLMs.  If the medical context (rules, association mappings, reference materials, etc.) does not exceed the input limits of the model, which at the time of writing is 3k and 25k words for ChatGPT and GPT4, respectively, associations and context can be included as input.  The process of manipulating LLM results through input content and structure is commonly referred to as prompt engineering \cite{white2023prompt}.  However, technical limitations aside, data policy, privacy, bias, and accuracy concerns associated with AI in medicine persist.  With limited information on the underlying data or model training process, it is not clear that third-party use of services like ChatGPT is consistent with FDA guidance \cite{us2021good} on the application of AI in clinical care.

LLMs possess a great capacity to analyze vast amounts of medical data, ranging from research papers and clinical trial results to electronic health records and patient narratives\cite{xue2023potential, dave2023chatgpt}.  By integrating these diverse, potentially multimodal\cite{li2023llava} data sources, these models can identify patterns, correlations, and insights that might have otherwise remained hidden.  With their ability to understand natural language, these AI-powered systems can process patient symptoms, medical histories, and test results to aid in diagnosing diseases more efficiently.  LLMs have demonstrated encouraging capabilities to generate \cite{zhou2023evaluation} and summarize \cite{temsah2023overview} medical reports, including radiology \cite{ma2023impressiongpt, biswas2023chatgpt, jeblick2022chatgpt} and pathology \cite{pathchatgpt, sinha2023applicability, brennan2023using} diagnostic reports.    

\begin{wrapfigure}{l}{8.2cm}
\vspace{-6mm}

\includegraphics[width=9cm]{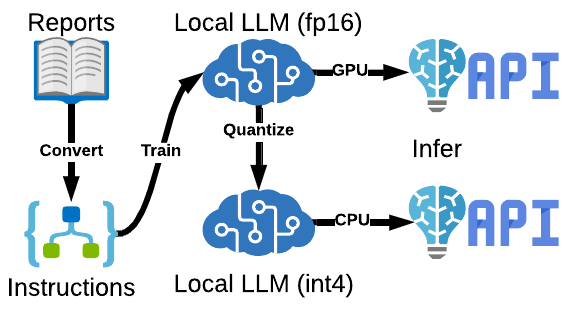}
\caption{local LLM High-level View}
\label{llmhigh}
\vspace{-4.75mm}

\end{wrapfigure} 

The theme of bigger is better continues to reign in the world of AI, especially as it pertains to language model data and parameter sizes.  Five short years ago, Google's BERT \cite{devlin2018bert} language transformers revolutionized deep learning for NLP tasks.  While large compared to vision models of the same generation, BERT-style models were publicly available, provided with permissive licenses, and rapidly incorporated into NLP pipelines.  BERT-style models are small enough to be fine-tuned for specific tasks, allowing the incorporation of medical and other data within the model itself.  The latest LLMs, such as GPT4, reportedly consist of trillions of parameters, are trained with trillions of input tokens, and reportedly cost hundreds of millions of dollars.  If publicly available, few institutions would have the expertise or capacity to infer GPT4-sized models, much less train them.  Fortunately, three months after the release of ChatGPT, Meta released LLaMA \cite{touvron2023llama}, and later LLaMA 2 \cite{touvron2023llama2}, which are foundational LLMs that are small enough to be trained, yet large enough to approach ChatGPT performance for many tasks.  Following the release of LLaMA, additional foundational models such as Falcon\cite{zxhang2023falcon} and MPT \cite{mosaicml2023introducing} were released.  Similar to previous community models such as BERT, these new foundational LLM models are provided in a range of sizes from 3 to 70 billion parameters.  Table \ref{modelsizes} provides the number of parameters, and Table \ref{modelmemory} lists vRAM requirements for common language models.  There are now tens of thousands\cite{llmleaderboard} of derivative LLMs trained for specific tasks, including the medical domain \cite{wu2023pmc}, which can benefit from both complex language reasoning and domain-specific training.  We will refer to LLMs that can be trained and operated without needing services, such as OpenAI and Google BART \cite{googlebart} as local LLMs. 

Using LLMs to extract machine-readable values is an area of research that has recently attracted significant attention.  This research aims to leverage the capabilities of LLMs to extract\cite{hu2023zero,wei2023zero} specific numerical or discrete information from unstructured text in a format that can be used by downstream computational pipelines.  Typical approaches to LLM structured data output include prompt engineering and post-processing, which can be applied to both services and local LLMs.  Most recently, projects such as Microsoft Guidance \cite{msguidance}, LangChain\cite{langchain}, and JsonFormer \cite{jsonformer} have emerged to manage the input structure, model interaction, and output structure of both online and local LLMs.  In addition, local LLMs can be fine-tuned to provide structured data in response to specific generative instructions, which can be combined with LLM data control software.

In this paper, we provide an approach to harness the language reasoning power of LLMs with the benefits of locally trained and operated models to perform complex, domain-specific tasks.  We will demonstrate our approach by extracting structured condition codes from pathology reports.  ChatGPT does not have sufficient medical context to report structured conditions from pathology reports, providing the response \textit{"I don't have the capability to perform specific queries to extract information like ICD codes from medical reports."}  Likewise, while BERT-style models work well for limited-sized text and frequently used condition codes, they lack the language processing capabilities to perform well across complicated unstructured data with high numbers of multi-label codes.  We test the efficacy of our local LLMs against BERT-style models that have been trained with pathology language data, and LongFormer \cite{beltagy2020longformer}, an extended context BERT-like model, both of which we fine-tuned for data extraction.  

\begin{table}
\vspace{-4mm}

\parbox{.45\linewidth}{
\centering
\begin{tabular}{ll}
\toprule
Model      & \# Parameters \\ \midrule
GPT 4      & 1.7t (reportedly)                        \\
GPT 3.5    & 175b \cite{openaichatgpt}                     \\
LLaMA     & 7b,13b,33b,65b \cite{vicunaparameters}              \\
Longformer & 149m \cite{bertparameters}                        \\
BERT-base  & 110m \cite{bertparameters}                        \\ \bottomrule
\end{tabular}
\caption{Comparison of model sizes}
\label{modelsizes}
}
\hfill
\parbox{.45\linewidth}{
\centering

\begin{tabular}{ll}
\toprule
Model      & vRAM \\ \midrule
LLaMA 7B      & 14GB                        \\
LLaMA 13B    & 27GB                     \\
LLaMA 33B     & 67GB              \\
LLaMA 65B  & 133GB                       \\ \bottomrule
\end{tabular}
\caption{LLaMA vRAM requirements}
\label{modelmemory}
}
\vspace{-8mm}
\end{table}

\section{Methods}
\label{methods}

This section will describe our process for curating LLM datasets,  model training and evaluation, quantization \cite{dettmers2022llm} approaches, and operational hosting of local LLM models.

\subsection{LLM Instruction Datasets}

We derived our dataset from over 150k uncurated surgical pathology reports containing gross descriptions, final written diagnoses, and ICD condition codes \cite{icd10} obtained from clinical workflows from the University of Kentucky.  ICD codes were used over other condition codes as they were available in new and historical reports.  Gross reports describe the characteristics of tissue specimens, and final reports describe the diagnosis based on microscopic review of tissues in conjunction with laboratory results and clinical notes.  A single case might contain many tissue specimens, which results in individual gross and final reports.  It is common practice in pathology reports to identify gross and final diagnosis specimen results within semi-structured templated text reports, with resulting specimen condition codes assigned to the entire case.  The result of this practice is that there is no direct association between case-reported condition codes and specimens.  It is common for there to be multiple condition codes per specimen, so conflicting codes can occur within a case.  For example, if one specimen is malignant and the other benign, the codes assigned to the case would conflict.  As a result of reporting practice, extracting condition codes on a specimen level is a complex NLP challenge.  Our motivation for this effort, beyond demonstrating the use of LLMs, is to better identify specimens and their related digital slides for multimodal and vision-based clinical AI efforts.               

We limited our dataset to cases with cancer-related codes, reducing the potential ICD label range from 70k to 3k.  We further eliminated cases that did not include condition codes, or a final report, reducing the dataset case count to 117k.  In order to test the performance of various model architectures and parameter sizes, we created three datasets: large (all data), small (10\% of large), and tiny (1\% of large).  For each dataset, code combinations that didn't appear at least 10 times were eliminated.  Training and test sets were generated with a 10\% code-stratified split. 
 The random sampling of cases in the reduced sets combined with the imposed code distribution requirements provides smaller datasets with more common codes.         

Given that the condition codes are reported on the case level, we concatenated gross and final reports into a single text input and assigned associated ICD codes as the output label.  Each model class and training system has its own format, which we will explain in the following sections.   

BERT and LongFormer models can be trained with the same datasets.  These datasets are most often CSV files, where the first column is the text input and the remaining columns are binary hot-encoded labels indicating label class, as shown in Table \ref{bertdataset}.

\begin{table}[ht]
\vspace{-4mm}
\centering
\begin{tabular}{@{}llll@{}}
\toprule
Input Text                                                            & code\_0 & code\_1 & code\_N \\ \midrule
biopsy basal cell carcinoma type tumor...     & 0       & 1       & 0       \\
lateral lesion and consists of tan soft tissue... & 1       & 0       & 0       \\
omentum omentectomy metastatic high grade carcinoma...      & 0       & 0       & 1       \\ \bottomrule
\end{tabular}
\caption{Example BERT and LongFormer training data format}
\label{bertdataset}
\vspace{-4mm}
\end{table}

LLMs are typically trained using an instruction-based format, where instructions, (optional) input, and model response are provided for one or more interactions in JSON format.  For each pathology case, we concatenate all text input into a single input field with the associated codes as the model response.  Each case is represented as a single conversation.  An example of an abbreviated case instruction is shown in Listing \ref{instructionformat}.    

\begin{lstlisting}[basicstyle=\fontsize{9}{10}\selectfont, caption=LLM Instruction JSON Format, label=instructionformat, style=codelistingstyle]
"id": "identity_0",
"conversations": [
    {
        "from": "human",
        "value": "right base of tongue invasive squamous cell carcinoma"
    },
    {
        "from": "gpt",
        "value": "C12\nC77"
    }
\end{lstlisting}

\subsection{Model Training}

As part of this effort, we trained over 100 models across multiple datasets, model architectures, sizes, and training configurations.  For each dataset (tiny, small, and large), we increased model size, where applicable, and training epochs, until the performance of the testing dataset diminished, which we discuss in detail in Section \ref{results}, \emph{Results}.  All training was conducted on a single server with 4XA100 80G GPUs\cite{choquette2021nvidia}.  For LLaMA 7B and 13B parameter models, the average training time was 25 minutes per epoch and two hours per epoch, respectively.  In the following sections, we describe the training process for each unique model architecture.      

\paragraph{BERT} and its successor transformer models are available in three forms 1) Foundational model, 2) Extended language model, and 3) Fine-tuned model.  Foundational models, as the name would suggest, are trained on a wide corpus of language, which provides a foundational model for fine-tuned tasks, such as code extraction.  

In areas where common language and words do not adequately represent the applied domain, unsupervised language modeling can be used to train a new model on domain-specific language.  For example, the popular BioBERT \cite{lee2020biobert} model, which was trained using biomedical text, has been shown to outperform the foundational BERT model for specific biomedical tasks.  Using example Hugging Face transformer language modeling code \cite{language_modeling}, we trained our own BERT-based language model using case notes as inputs.  Except for the removal of condition code columns, the training data is identical to the format shown in Table \ref{bertdataset}.         

All BERT models were fine-tuned using example Hugging Face transformer training code \cite{bert_trainer}.

\paragraph{LongFormer} is a BERT-like model that makes use of a sliding window and sparse global attention, which allows for an increased maximum input token size of 4096 compared to 512 for BERT.  While the majority of gross or diagnostic reports would not exceed the capacity of BERT models, the concatenation of report types across all specimens in the case could easily exceed the 512-token limit.  LongFormer models, which provide twice the input token size of our local LLM (2048), allow us to test the impacts of maximum token size on BERT-style model performance.

No language modeling was performed with LongFormer models, and all models were fine-tuned using example Hugging Face LongFormer transformer training code \cite{longformer_trainer}.

\paragraph{LLaMA-Based LLMs} are by far the most popular local LLM variants.  Models can vary based on training data, model size, model resolution, extended context size\cite{chen2023extending}, and numerous training techniques such as LoRA\cite{hu2021lora} and FlashAttention\cite{dao2023flashattention}.  Research associated with local LLMs is developing at a very rapid pace, with new models and techniques being introduced daily.  The result of such rapid development is that not all features are supported by all training and inference systems.  Fortunately, support has coalesced around several projects that provide a framework for various models and experimental training techniques.  We make use of one such project named FastChat\cite{zheng2023judging}, an open platform for training, serving, and evaluating large language models.  The FastChat team released the popular LLaMA-based LLM Vicuna.  Following the Vicuna training code described by the FastChat team, we trained our LLMs using our pathology case data in instruction format, as shown in Listing \ref{instructionformat}.  We trained both 7B and 13B parameter LLaMA models across our three datasets.  In all cases, our LLaMA-based models were trained with half-precision (fp16).

\subsection{Local LLM Hosting}

As previously noted in Table \ref{modelsizes}, the sizes of foundational language models have grown significantly since the release of BERT.  As model sizes increase, model-level parallelism must be used to spread model layers across multiple GPUs and servers.  In addition, model checkpoints themselves can be hundreds of gigabytes in size, resulting in transfer and load latency on model inference.  The development of inference services that implement the latest models, techniques, and optimize resource utilization is an active area of research.  We make use of vLLM \cite{vllm}, an open platform that supports numerous model types, extensions, and resource optimization.  vLLM, and other inference platforms, provide API services, allowing users to decouple inference services from applications.  In addition, vLLM includes an OpenAI-compatible API allowing users to seamlessly compare ChatGPT/GPT4 with local LLMs results.       

Unless otherwise noted, all local LLM performance testing was conducted using vLLMs OpenAI-compatible API.   

\paragraph{Generative Pre-trained Transformer Quantization} (GPTQ)\cite{frantar2022gptq} is a technique that is used to reduce the GPU memory requirements by lowering the precision of model weights and activations.  To match the resolution of the foundational LLaMA models and to reduce resource requirements, local LLMs are commonly trained at half- (fp16) or quarter-precision (int8).  However, even at half-precision, the GPU memory requirements are significant and can exceed the capacity of the largest single GPUs, as shown in Table \ref{modelmemory}. 

\paragraph{Quantization for CPUs} has become extremely popular as LLM model sizes and associated resource requirements increase.  Using CPU-focused libraries, such as GGML\cite{ggml}, models can be further quantized to even lower precision (int4, int3, int2).  High levels of quantization can drastically reduce resource requirements and increase inference speed, allowing LLMs to be run directly on CPUs.  As with model size, the performance impacts of precision reduction are highly dependent on the workload.  Quantization can occur post-training, allowing a single model to be trained and reduced to various quantization levels for evaluation.       
Similar to vLLM, LLaMA.cpp \cite{llamacpp} is an open platform that focuses on the support of GGML quantized models on CPUs.  LLaMA.cpp provides tools to quantize pre-trained models and supports bindings for common languages such as Python, Go, Node.js, .Net, and others.  The LLaMA.cpp Python \cite{llamacpppython} project provides an OpenAI-compatible API, which we use to evaluate quantized local LLMs where indicated.

\section{Results}
\label{results}

Seven different model architectures were tested on the three dataset sizes (tiny, small, large). This includes four separate BERT models: BERT-base-uncased, BioClinicalBERT, PathologyBERT \cite{santos2022pathologybert}, and UKPathBERT.  BERT-base-uncased is the original foundational BERT model, BioClinicalBERT is trained on biomedical and clinical text, PathologyBERT is trained on pathology reports that are external to our institution, and UKPathBERT is our own BERT-base-uncased language model trained on our own pathology report dataset.

Additionally, the BERT-like Longformer model with an increased input context size was trained.  The performance of these BERT-style models serves as benchmarks and evidence for the complexity of our language tasks.  

Finally, LLaMA 7b and 13b parameter models were trained using the same datasets in an instruction-based format, which we will refer to as Path-LLaMA.  Unlike most other generative LLMs, our intended output is a structured set of condition codes.  As previously mentioned, we experimented with pre- and post-processing techniques to ensure structured output.  We achieved the best results by ordering condition codes into alphabetical lists separated by line breaks.  With the exception of single epoch training of the Path-LLaMA 7b model, deviation (hallucination) from the intended format was not experienced.  The stability of the structured output allowed us to statistically evaluate model results as we would other non-generative models.           

In both generative (LLM) and BERT-style transformer model cases multi-label classification results will be evaluated the same.  Accuracy (ACC) refers to the frequency of explicitly correct predicted labels. For example, if a particular case has two labels assigned to it and the model correctly guesses one of them, the accuracy is 0\% for that case. Because of this strict method, accuracy is somewhat low compared to the other performance metrics.  The AUC (Area Under the ROC Curve) is calculated for each possible class, and the macro (unweighted) average is taken. This was performed using the sklearn metrics \cite{sklearnmetrics} package. In the context of multilabel classification, the AUC represents how likely each class is to be labeled correctly. Therefore, similarly to binary classification, an AUC below 0.5 represents that the model performs worse than random chance on average.  Similarly, precision, recall, and F1 score are calculated for each class and macro averaged together to produce a final result, using the sklearn metrics package's classification report function. With multilabel classification, precision measures the proportion of correct predictions, while recall measures the proportion of instances that received correct classifications, and the F1 score averages these together.

The best results of any architecture were achieved with the LLaMA-based LLM, as seen in Table \ref{overallresults}, which shows the overall model performance results, averaged across all datasets and parameter settings, such as number of epochs trained. 
\begin{table}[h]
\vspace{-4mm}
\centering
\begin{tabular}{llllll}
                        \toprule
                         & Accuracy & AUC      & Precision & Recall   & F1 \\
                         \midrule
Path-LLaMA 13b               & 0.748 & 0.816 & 0.779  & 0.777  & 0.775 \\
Path-LLaMA 7b                & 0.647 & 0.763 & 0.68  & 0.674  & 0.674 \\
UKPathBert   & 0.058 & 0.506 & 0.059  & 0.059  & 0.059 \\
PathologyBERT           & 0.057 & 0.502 & 0.059  & 0.059  & 0.059 \\
BioClinicalBERT       & 0.053 & 0.507 & 0.055  & 0.054  & 0.055 \\
BERT-base-uncased & 0.036 & 0.498 & 0.04  & 0.042  & 0.04 \\
Longformer 149m          & 0.001 & 0.5 & 0.063  & 0.42  & 0.103 \\
\bottomrule
\end{tabular}
\caption{Average performance of each model on all datasets}
\label{overallresults}
\vspace{-4mm}
\end{table}

The largest LLaMA model, with 13 billion parameters, performed the best on average. Both Path-LLaMA models performed significantly better than any other model. The BERT transformers performed poorly on average, but the versions that were trained specifically on pathology-related text outperformed the basic model. The Longformer had better recall than the BERT models because it tends to predict many different codes, meaning it has a higher chance of guessing correctly. However, this brings down the precision and accuracy of this model because many of its guesses are wrong.

LLaMA-based models outperform BERT-style models across all evaluation metrics.  As expected, larger parameter models tend to outperform smaller models, and models trained within a specific domain, outperform those that are not.  In the remainder of this section, we go into more detailed evaluations of model size, numbers of epochs, dataset size, and other potential performance factors.

\subsection{Model Size}

Two most commonly used sizes of LLaMA models 7b and 13b were tested to determine the impact of parameter size on performance.  In testing, we observed very similar inference performance of 0.3-0.4 seconds per case between 7b and 13b models using fp16.  We attribute this to our multi-gpu test system, which is less utilized with the 7b model, and other overhead of the decoupled API interface.  We also tested GGML int4 quantized versions of 7b models, which for results were nearly identical to their fp16 counter parts, but with an inference time of 7.5 seconds per case.  Despite the lower precision, CPU-based inference resulted in significantly longer inference times.            

As seen in Table \ref{overallresults}, the larger model performed better on average. However, when compared only to the large datasets, their performance was very similar. Both achieved an F1 score of 0.785, while the 13b model obtained a slightly higher accuracy of 0.742 compared to the 7b model's 0.737. This seems to demonstrate that the increase in size had little effect on performance when compared to the largest dataset.

\subsection{Number of Epochs}
Each model architecture was trained on a range of epochs. The number of epochs tested for each was dependent on two things: model training time (dataset and parameter sizes) and how many epochs it took before the results on the training set no longer improved. The average F1 for each model and the number of epochs are given in Table \ref{epochs}.

\begin{table}[h]
\vspace{-4mm}
\centering
\begin{tabular}{llllllll}
\toprule
                         & 1     & 3     & 6     & 12    & 24    & 48    & 96    \\ \midrule
Path-LLaMA 13b               & 0.749 & 0.767 & \textbf{0.80} &       &       &       &       \\
Path-LLaMA 7b                & 0.486 & 0.586 & 0.761 & \textbf{0.825} & 0.759 &       &       \\
UKPathBERT   & 0.002 & 0.008 & 0.032 & 0.041 & 0.148 & \textbf{0.2}   & \textbf{0.2}   \\
PathologyBERT           & 0.004 & 0.006 & 0.055 & 0.037 & 0.117 & \textbf{0.267} & 0.133 \\
BioClinicalBERT         & 0.004 & 0.007 & 0.009 & 0.059 & 0.118 & \textbf{0.4}   &       \\
BERT-base-uncased & 0.015 & 0.007 & 0.007 & 0.016 & 0.088 & \textbf{0.2}   & 0.133 \\
Longformer               & 0.081 & 0.075 & 0.072 & \textbf{0.229} & 0.219 &       &       \\ \bottomrule
\end{tabular}
\caption{F1 of each model for each number of epochs tested}
\label{epochs}
\vspace{-4mm}
\end{table}

This table shows that the number of epochs during training can have a significant impact on the results of the model.  In each case, at least six epochs were required to train the best model, in some cases significantly more.  Optimal epoch count is very much experimental in practice, as it is highly dependent on the dataset, model parameter size, and other training parameters.   

\subsection{Dataset Size}

\begin{wraptable}{r}{0.5\textwidth}
\vspace{-8mm}

\begin{tabular}{llll}
\toprule
                         & Tiny  & Small & Large \\ \midrule
Path-LLaMA 13b               & 0.778 & 0.761 & \textbf{0.785} \\
Path-LLaMA 7b                & 0.641 & 0.764 & \textbf{0.783} \\
UKPathBERT   & \textbf{0.114} & 0.014 & 0.018 \\
PathologyBERT           & \textbf{0.114} & 0.011 & 0.021 \\
BioClinicalBERT         & \textbf{0.105} & 0.012 & 0.019 \\
BERT-base-uncased & \textbf{0.073} & 0.012 & 0.018 \\
Longformer               & \textbf{0.206} & 0.025 & 0.009 \\ \bottomrule
\end{tabular}
\caption{F1 of each model on each dataset}
\label{datasets}
\vspace{-4mm}

\end{wraptable}

The average number of words per pathology case was approximately 650, so assuming token counts are 1.25X larger than words, our largest dataset contained over 80M tokens from 100k cases.  As previously mentioned, larger datasets include a wider range of condition codes, so in this context, a larger dataset does not necessarily guarantee better performance.  The performance of each model on each dataset size is shown in Table \ref{datasets}.

BERT and Longformer models all performed best on the smallest dataset, while the LLaMA models performed best on the largest. The smaller dataset is an easier classification problem, with fewer possible class labels and examples, but the larger dataset has more complex data to train from. This seems to further reinforce the superiority of LLaMA compared to the other models. When the dataset is large, the other models fail, while LLaMA only improves with more data, demonstrating its improved capability to learn and correctly classify condition codes compared to the other models.

\subsection{Other Result Factors}
In this section, we cover additional model performance factors, such as the length of input text and the frequency of classification codes in our dataset.

The length of the input description for each sample was paired and analyzed with how often that sample was predicted correctly for each model. This was done to determine if, for example, longer descriptions allowed the model to understand the text better and classify the correct code more often. However, it was found that there was no significant correlation between the length of the description and how often that sample was correctly predicted.  We speculate that the complexity of language far outweighed the size of the input context window, as indicated by LongFormer performance. 

Certain classification codes were far more frequent in the dataset than others. This was especially true for the tiny and small datasets, which might have only ten examples of specific code combinations. The frequency of each code in the dataset was analyzed along with what percentage of the time that code was correctly predicted by the models. Unsurprisingly, it was found that the most common codes were predicted correctly more often when compared to the less common classification codes.  Likewise, smaller models performed better with a limited range of codes.

\section{Conclusion}
\label{conclusion}

In this paper, we described the end-to-end process of training, evaluating, and deploying a local LLM to perform complex NLP tasks and provide structured output.  We analyzed model performance across parameters and data size along with data complexity.  We compared these results with BERT-style models trained on the same data.  The results of this effort provide overwhelming evidence that local LLMs can outperform smaller NLP models that have been trained with domain knowledge.  In addition, we demonstrate that while more latent, LLMs can be deployed without GPUs.  While we make no claims that local LLMs provide comparable language processing capabilities to ChatGPT and its successors, technical and policy limitations make local LLMs actionable alternatives to commercial model services. We have also shown that accurate models (such as LLaMA 7b) can be made usable on reasonable CPU/GPU hardware with minimally increased overhead.

In future efforts, we aim to explore newer and larger models, such as LLaMA 2 and Falcon.  We would like to further explore the impact of LLM context size and post-training context extension on model performance.  Finally, we aim to explore the structure of instruction and input training data on model results. 

With the exception of the identified example dataset, code and instructions to recreate this work can be found in the following repository: \url{https://github.com/innovationcore/LocalLLMStructured}

\section{Acknowledgements}
\label{acknowledgements}

The project described was supported by the University of Kentucky Institute for Biomedical Informatics; Department of Pathology and Labratory Medicine; and the Center for Clinical and Translational Sciences through NIH National Center for Advancing Translational Sciences through grant number UL1TR001998.  The content is solely the responsibility of the authors and does not necessarily represent the official views of the NIH.

\bibliographystyle{ws-procs11x85}
\bibliography{0_citations}

\end{document}